\documentclass[12pt,a4paper,final]{iopart}

\usepackage{iopams}  
\usepackage{graphicx}
\usepackage[breaklinks=true,colorlinks=true,linkcolor=blue,urlcolor=blue,citecolor=blue]{hyperref}
\usepackage{dcolumn}
\usepackage{bm}

\expandafter\let\csname equation*\endcsname\relax
\expandafter\let\csname endequation*\endcsname\relax
\usepackage{amsmath}
\usepackage{epstopdf}

\usepackage[usenames, dvipsnames]{color}

\usepackage{array}
\usepackage{float}
\usepackage{multirow}

\begin{document}


\title[Patterns in the English Language]{Patterns in the English Language: Phonological Networks, Percolation and Assembly Models. }

\author[cor1]{Massimo Stella}
\address{Institute for Complex Systems Simulation, University of Southampton, Southampton, UK}
\ead{massimo.stella@inbox.com}

\author{Markus Brede}%
\address{Institute for Complex Systems Simulation, University of Southampton, Southampton, UK}

\date{October 15, 2014}

\begin{abstract}
In this paper we provide a quantitative framework for the study of phonological networks (PNs) for the English language by carrying out principled comparisons to null models, either based on site percolation, randomization techniques, or network growth models. In contrast to previous work, we mainly focus on null models that reproduce lower order characteristics of the empirical data. We find that artificial networks matching connectivity properties of the English PN are exceedingly rare: this leads to the hypothesis that the word repertoire might have been assembled over time by preferentially introducing new words which are small modifications of old words. Our null models are able to explain the ``power-law-like'' part of the degree distributions and generally retrieve qualitative features of the PN such as high clustering, high assortativity coefficient, and small-world characteristics. However, the detailed comparison to expectations from null models also points out significant differences, suggesting the presence of additional constraints in word assembly. Key constraints we identify are the avoidance of large degrees, the avoidance of triadic closure, and the avoidance of large non-percolating clusters.

\end{abstract}

\maketitle


\section{Introduction}
Complex networks can be used to model pairwise relationships between entities. In the last twenty years this approach has proved fruitful for gaining insights into a large number of complex systems, among them financial markets, social systems, and technological and biological webs \cite{newman2010,barabasi1999,amaral2000,ravasz2003,peel2014,antonioni2012,hagen2012}. The success of the network approach was made possible by the availability of an unprecedented amount of data, allowing for the analysis and discovery of common features in at first sight very different systems through the lens of the network paradigm \cite{newman2010}.

More recently, complex networks have also been investigated in the field of the cognitive sciences. Prominent examples are the modelling of structural patterns of connectivity in the human brain \cite{bullmore2009,sporns2004} or studies of cognitive processes, such as free-word associations \cite{dall2006}. Furthermore, complex networks represent a powerful quantitative tool for modelling the \textit{human mental lexicon} or HML \cite{motter2002,steyvers2005,sigman2002,i2001small}. The HML is an abstract representation of how words and their relative concepts are stored within the human brain \cite{aitchison2012}. One can imagine the HML as a huge database where words are stored together with additional information, and in which words are related by specific correlations which ease navigation, e.g. words can be opposites or synonyms, might be pronounced in similar or dissimilar ways, be related to the same context area, etc. Following a connectionist approach \cite{aitchison2012,griffiths2007} the HML can be interpreted as the representation of the biological patterns of synchrony/asynchrony among the $10^{10}$ neurons and $10^{14}$ synapses of the human brain \cite{baronchelli2013,bullmore2009}.

Ultimately, the various relationships present in the HML could only be adequately captured by a multi-layered network \cite{sole2014}. However, various layers have already been analysed in isolation, among them semantic networks (i.e. networks where nodes represent words and edges represent semantic relationships) which have been found to be small-worlds \cite{griffiths2007,i2001small,sigman2002,steyvers2005}, a characteristic which has been related to certain robustness properties of the organisation of human language \cite{motter2002}. Interestingly, the small world property in semantic networks depends strongly on ambiguity in language \cite{sole2014,de2008word}, which might be explainable by least effort principles applied to communication  \cite{icancho2003}.

By constructing \textit{phonological networks} Vitevitch \cite{vitevitch2008} also applied the network paradigm to modelling phonological patterns in English. In this construction nodes represent phonological transcriptions of words and edges indicate phonological similarity based on a similarity metric established in the field (cf., the phonological neighbourhood density \cite{cutler1977,luce1998,sadat2014}). The main motivation for a complexity approach to modelling the phonological structure of human language via network tools is that traditional psycholinguistic research has focused on local scale analyses to identify the role played by given lexical characteristics (e.g. word frequency, age of acquisition, word length) in determining the accuracy and speed  of retrieval of a given word from the mental lexicon \cite{luce1998,vitevitch1997}. Although this approach has been valuable, a globally detailed understanding of structural patterns in the mental lexicon is still an open research question \cite{vitevitch2014} -- a prime motivation for this study. For this purpose, the network approach is a suitable choice since it provides us with an established set of measures and tools to quantify local and global structural patterns.

Vitevitch's first analysis \cite{vitevitch2008} found the phonological network for $20,000$ English words to be disconnected, comprised of a giant component of almost $10^4$ words, a variety of smaller-sized components (termed ``linguistic islands''), and a very large number of isolated nodes (termed ``hermit words''). Furthermore, the giant component exhibits the small-word property combined with high cliquishness, a rather high level of assortative mixing by degree, and a degree distribution that has been described as a power-law with cut-off  \cite{vitevitch2012,chan2010,arbesman2010}. These results have been confirmed for phonological networks constructed for various other languages, such as  Spanish, Mandarin, Hawaiian, and Basque \cite{arbesman2010,arbesman2010b}. Building on these insights, in \cite{siew2013} it was shown that the giant component of the English PN exhibits also a rich community structure, in which large communities are preferentially composed of short, frequent and highly connected words with low age of acquisition ratings. These findings strongly suggest that larger communities may actually be the first to form during the assembly of the mental lexicon, and therefore they may be essential in determining the final structure of the PN \cite{siew2013}. This hypothesis is supported by the fact that phonological neighbours play a role in predicting the order of acquisition of nouns \cite{hills2010}, and by the empirical evidence that late talkers tend to acquire semantically novel words relative to known words in a way that significantly alters the small-world property of the English PN of normal speakers \cite{beckage2011}.

Analysing phonological networks it is important to realise that nodes (i.e. words) correspond to sequences of symbols (i.e. phonemes). Then the set of all possible combinations of symbols (together with the phonetic similarity metric) defines a space, of which the actual word repertoire is a subset. In the language of percolation \cite{stauffer1991} one might speak of occupied and empty nodes, corresponding to words that are actually present in language and hypothetically possible but not realized words. As is the case for networks in more conventional Euclidean spaces \cite{newman2003} the topology of the underlying space also constrains the organisation of phonetic networks. To some extent this has been realised by Grunenfelder and Pisoni  \cite{gruenenfelder2009} who carried out percolation style experiments similar to those of Mandelbrot \cite{mandelbrot1953} to generate phonological pseudolexica, but restricted the study to very short words composed of between two and five phonemes. Corresponding phonological networks were found to retrieve some qualitative characteristics of the English PN \cite{gruenenfelder2009} such as high clustering and strong assortative mixing by degree and the authors suggested that peculiarities of this network, as stated by \cite{vitevitch2008}, might be an artifact of the construction method. Whilst it is certainly true that the topology of the underlying space biases characteristics of the PN in the ways described by Grunenfelder and Pisoni, lack of quantitative agreement and the use of word length distributions that ignore longer (less connected) words make final conclusions difficult. Further, comparisons of higher order network statistics (as clustering or assortativeness) are not necessarily compelling when lower order characteristics (such as the number of links or sizes of components or degree sequences) differ markedly. In contrast to \cite{gruenenfelder2009}, a study across different languages \cite{arbesman2010} reiterated the original point of Vitevich. As a result of these contradicting findings the main point: ``Which characteristics of PNs are archetypical for organisations of words in language and which are mere artifacts of the construction method'' remains unresolved.

In this paper we develop a series of null models to carry out a principled analysis of the phonological network for English. Since our study of the English PN is based on a different database than previous work, we start by briefly reviewing some network properties. We then proceed by comparing the English PN to networks obtained from randomized sets of words. This naturally leads us to consider various types of percolation-style experiments that increasingly respect phonetic constraints. Comparisons to the English PN reveal significant differences in link counts and component distributions between lexicons of real words and pseudolexica. These differences hint at the presence of constraints on clustering and maximum degree in word assembly while pointing out that the power-law like part of the degree distributions observed in \cite{arbesman2010} appears as a natural consequence of the embedding space. We then refine these insights by using Monte Carlo Markov Chain (MCMC) like techniques to generate ensembles of words whose network representations have the same link counts as the original data set. The analysis reveals that differences in the sizes of giant components are particularly significant, suggesting the possibility of word assembly mechanisms that proceed by generating new words through small modifications of already existing words.

The question how likely it is to assemble pseudolexica that match link counts and component sizes of the English lexicon arises naturally. We next address this question by introducing various types of attachment models. Extending these models allows to construct ensembles of networks which match essential lower order statistics of the English PN. Hence a quantitative assessment of peculiarities of phonetic word organisation through network analysis becomes possible. The paper concludes with a discussion of these results in the light of constraints that might have shaped the assembly of the human mental lexicon.

\section{Network construction and analysis}

The construction of the PN for English adopted in this paper is based on roughly $30,000$ English words using the database from Wolfram Research, a curated repository mainly based on Princeton University Cognitive Science Laboratory "WordNet 3.0." \cite{miller1995} and on Oxford University Computing Service, British National Corpus, version 3 \cite{leech1992}. Phonetic transcriptions in this database are given using the International Phonetic Alphabet (IPA). Before constructing the networks we remove any \textit{supra-segmental} feature such as stress marks or accents and also remove all homophones, i.e. words with identical phonological transcriptions. Network construction then proceeds by associating the remaining words with nodes and connecting them whenever the respective words have edit distance one. It is worthwhile pointing out that the use of edit distance one to define connections is a to some extent arbitrary choice. Other choices are possible, but in the present study we follow earlier work \cite{luce1998,vitevitch1997,chan2010} which has related this choice to other measures established in psycholinguistics.

Some network statistics for the resulting network are summarised in table \ref{tab.1}. Comparison with the networks \cite{vitevitch2008,gruenenfelder2009,siew2013,vitevitch2012} based on the smaller $20.000$ words Hoosier Mental Lexicon (HML) \cite{aitchison2012} gives good quantitative agreement. For instance, in our database the giant component comprises $33\%$ of the nodes ($34\%$ for the HML), the clustering coefficient \cite{newman2010} $CC$ is $0.21$ (compared to $0.22$) and the assortativity coefficient $a$ \cite{newman2010} is $0.70$ (compared to $0.67$ for the HML). As expected mean geodesic path lengths are larger for our larger lexicon, i.e. $d=7.71$ (whereas $d=6.08$  for the HML). As already observed in previous works \cite{vitevitch2008,gruenenfelder2009,arbesman2010}, the giant component of the PN for English has the small-world property, i.e. when compared to similar size random graphs, it exhibits a higher clustering coefficient and similarly low average shortest path length \cite{newman2010}. Furthermore, on average, each linguistic island contains $2.49\pm0.04$ words, in agreement with the $\sim2.52$ estimate from \cite{gruenenfelder2009}. Also, the degree distribution of the giant component follows a power-law like behavior with a cut-off, similar to the analysis of \cite{arbesman2010} for English and several other languages.

Altogether, our larger dataset is able to closely reproduce features of the English PN as constructed from smaller databases, and we find similar macro (degree distribution, assortative mixing by degree, average clustering coefficient, average path length) and micro (node degree and local clustering coefficient) characteristics as previous analyses.

\section{Randomization experiments}

In order to investigate the features of the empirical English phonological network we next present a series of null models. The agenda for those is as follows. We first assume that system sizes are fixed and obtain reference models via randomization of the existing dataset. The first step leads us to percolation-based approaches presented in section \ref{sec:percolation}. These experiments could be seen as the generation of artificial word repositories generated from the original one through a shuffling procedure that preserves the word count. The approach can be refined by using MCMC like randomization procedures \cite{snijders2002} which allow the generation of randomized ensembles that preserve additional constraints. In section \ref{sec:mcmc} we apply this idea to generate ensembles of artificial words whose corresponding networks have exactly the same number of connections as the original data. 

The randomization approach is reminiscent of Exponential Random Graph Modelling (ERGM), which aims to identify the Hamiltonian that defines a network ensemble a given empirical network might be an example of \cite{newman2010}. ERGM is particularly useful when one wants to create network models matching the features of empirical networks as closely as possible without specifying the details of the underlying network formation process \cite{fronczak2012}. In the last decade ERGM has been applied widely in social network analysis \cite{robins2007}, particularly for the analysis of reciprocity and transitivity in friendship and contagion networks. This method typically relies on the use of computationally demanding MCMC simulations combined with regression techniques. Applying ERGM models to the analysis of phonological networks could give interesting additional insights, but (i) computational costs and (ii) open theoretical questions on how to include constraints from spatial embedding (which has only been considered in Euclidean space quite recently \cite{daraganova2012}) and (iii) phoneme statistics make the randomization approach the preferred choice for this study. 

Following on from the randomization analysis in Sec. \ref{Growth} will consider if the English PN can be modelled via processes of word assembly over time. In the remainder of the paper
 we will refer to a table that summarizes network statistics for all reference models we present, cf. Table \ref{tab.1}. To avoid confusion we also provide a table that gives an overview over all models, cf. Table \ref{tab.0}.

\begin {table*}[tbp]
\begin{center}
\resizebox{\columnwidth}{!}{
\begin{tabular}{| c | c | p{9 cm} |}
  \hline \hline
  Notation & Type & Brief Explanation \\
  \hline
  \hline
type 0 & Percolation Experiment & Percolation-like experiment with uniform random sampling of phonemes, which generates a pseudolexicon with the same word length distribution as the real data. \\
\hline 
type 1 & Percolation Experiment & The same as in type 0 but with additional constraints on the phoneme frequencies.\\
\hline 
type 2 & Percolation Experiment & The same as in type 1 but with additional constraints on the phoneme-phoneme correlations.\\
\hline
MCMC & Monte Carlo Markov Chain & Ensemble of randomized networks which are obtained from the English PN via word shuffling that enforces the empirical word length distribution and realistic phoneme statistics as in type 2 and additionally enforces a link constraint. The method generates an ensemble of networks with the same number of connections as the English PN.\\
\hline
grow gc (r) & Growing Network & The network is grown over time. At each timestep a new word is generated at random (as in type 2 experiments). If it does not receive any connection, then it is discarded with probability $f$.  Words with degree larger than $k_{max}$ are additionally suppressed.\\
\hline 
grow gc (o) & Growing Network & The same as in grow gc (r) but here shorter words are generated first.\\
\hline 
($W_{C}$, $m_{C}$), $f$ & Core-Periphery Models & The same as in grow gc (r) experiments but with an additional chance of rejection for words shorter than $W_C$ with less than $m_{C}$ links. These experiments enforce the generation of a network core, which is composed of preferentially short and highly connected words.\\
\hline 
$(W_{C},m_{C})$, $\epsilon$, $f$ & Core-Periphery Models & The same as in ($W_{C}$, $m_{C}$), $f$ experiments but with an
additional rejection probability $\epsilon$ of discarding words connecting to the giant component.\\
\hline
\end{tabular}
}
\caption{Summary of all the null models implemented in this paper, as reported in Table \ref{tab.1}.}
\label{tab.0}
\end{center}
\end{table*}

\subsection{Percolation}
\label{sec:percolation}

\begin{figure}[tbp]
 \begin{center}
\includegraphics[width=.6\textwidth]{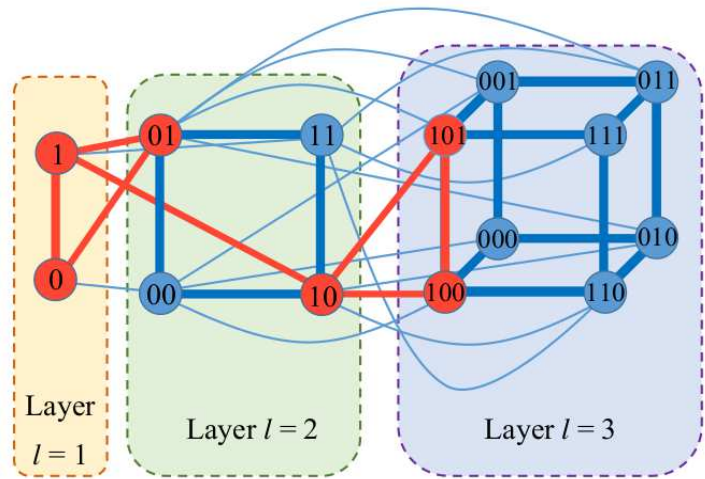}
 \caption {Visualisation of a substrate graph with a binary phonetic alphabet $\cal{P}$ $=\{0,1\}.$ In this case, each layer is represented as a hypercube. Red nodes represent the actual words in a fictional binary language. Red links connect phonologically similar actual words. The other connections between layers have been omitted for a better visualisation.}
\label{fig.n0}
\end{center}
 \end{figure}

Let us introduce the set of all possible phoneme sequences $S=\cup_l L_l$, where $L_l=\{w_l\}$ is the set of all possible words $w_l=\{s_0,...,s_l\}$, $s_i\in \cal P$ of length $l$ and the set $\cal P$ is the set of all possible phonemes of a given language. For example, for our dataset of English words, we have $|{\cal P}|=36$ phonemes and words up to length $l=21$. Within a set $L_l$ distances between words can be measured by the conventional Hamming distance, between words in different sets distances can be determined by the minimum number of phoneme additions, deletions are substitutions required to transform one word into another, the so-called edit distance $d_E(\cdot,\cdot)$, a generalisation of the Hamming distance \cite{ristad1998}. By identifying nodes with possible phoneme sequences, i.e. the set $S$, and connecting nodes whenever their associated words have edit distance one, a substrate graph is defined, of which phonological networks are a subset. For a better understanding of this substrate graph it is useful to visualise it as a stacked set of \textit{layers} of graphs composed of words of the same length, cf. Figure \ref{fig.n0}. In this way one can naturally distinguish between \textit{intra-} and \textit{inter-layer} connections which encapsulate additional information about phoneme organisation, on top of more conventional network measures.

\begin{figure}[tbp]
 \begin{center}
\includegraphics[width=.65\textwidth]{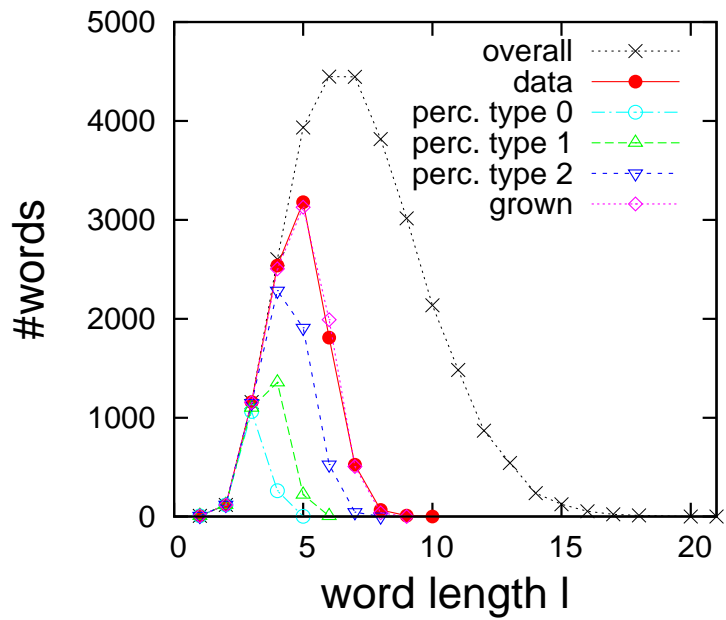}
 \caption {Word length distribution for English and distribution of word lengths in the giant component. Data for the word length distribution in the giant component are compared for the dataset of phonetic transcriptions of English words, type 0, 1, and 2 percolation experiments and the growth experiments introduced in Sec. \ref{Growth} for $k_\textrm{max}=25$ and $f=0.75$. Data points represent averages over at least 10 configurations.}
\label{fig.n1}
\end{center}
 \end{figure}

The organisation of substrate graphs in such layers leads to an evident conclusion. Since the number of all possible words of given length grows exponentially $|L_l|=|{\cal P}|^l$, whereas the number of actual words of given length grows markedly less rapidly occupation densities of layers decrease exponentially with increasing word length $l$. Also, coordination numbers of nodes in $L_l$ are given by $\kappa_l=(|{\cal P}|-1)l$. Using the Bethe approximation as in \cite{stauffer1991} as a rough estimate of percolation thresholds of individual layers, one thus expects to find giant components only in layers made up of shorter words, for which the percolation density threshold is exceeded. This points to the importance of the word length distribution $H(l)$ in determining properties of phoneme networks. Word length distributions with a bias for shorter words will naturally induce larger component sizes than word length distributions that account for relatively more long words. Hence, comparisons between pseudolexica and real datasets are only sensible if the same word length distribution is used. Figure \ref{fig.n1} gives the word length distribution for our dataset of phonetic transcriptions of English words.

\begin {table*}[tbp]
\begin{center}
\resizebox{\columnwidth}{!}{
\begin{tabular}{| l | c | c | c | c | c | c | c | c | c | c |}
  \hline \hline
  experiment & $L$ & $L_0$ & $lr$ & $gc$ & $k_\text{max}$ & $CC$ & $a$ & $d$ & $d_\text{max}$ \\
  \hline
  \hline
  English           & $38342$ & $34896$ & $2.46$  &  $9412$ & $44$  & $0.207$ & $0.707$ & $7.71$ & $33$ \\
  \hline \hline
    MCMC              & $38342$ & $37845\pm 10$ & $2.21\pm 0.01$ &
$7260\pm 20$ & $88\pm 1$ & $0.304\pm 0.001$ & $0.41\pm 0.01$ &
$5.47\pm 0.01$ & $20.3\pm 0.4$ \\
  \hline \hline
  type 0         & $2840\pm 50$  & $2490\pm 50$ & $3.39\pm0.01$   &  $1440\pm40$ & $18.4\pm0.8$ & $0.16\pm0.01$  & $0.59\pm0.01$  & $7.63\pm0.01$ & $22.5\pm0.5$ \\
  type 1         & $8370\pm30$ & $8210\pm30$ & $2.59\pm0.01$  &  $2808\pm20$ & $46.8\pm0.7$ & $0.220\pm0.001$  & $0.46\pm0.01$  & $5.45\pm0.01$ & $16.1\pm0.3$\\
  type 2         & $20420\pm30$ & $19820\pm30$ & $1.95\pm0.01$  &  $6020\pm20$ & $52.7\pm0.8$ & $0.248\pm0.001$  & $0.45\pm0.01$  & $5.95\pm0.01$ & $18.9\pm0.3$ \\
  \hline \hline
  grow gc (r)    & $28890\pm20$ & $27740\pm20$ & $1.68\pm0.01$  &  $9390\pm20$ & $43.1\pm0.7$ & $0.259\pm0.001$  & $0.46\pm0.01$  & $6.53\pm0.01$ & $18.8\pm0.3$ \\
  grow gc (o)    & $27260\pm20$ & $27110\pm20$ & $1.64\pm0.01$  &  $9380\pm20$ & $42.2\pm0.9$ & $0.254\pm0.001$  & $0.47\pm0.01$  & $6.59\pm0.01$ & $19.0\pm0.4$ \\
  \hline  \hline
  $(5,4.2), f=0.76$      & $35580\pm40$ & $34900\pm40$ & $2.17\pm0.01$  &  $9380\pm40$ & $43.2\pm0.0.9$ & $0.258\pm0.001$  & $0.48\pm0.01$  & $6.35\pm0.01$ & $23.8\pm0.4$ \\
  \hline
  $(5, 4.4)$, $\epsilon=0.82$, $f=0.962$   & $38320\pm30$ & $34860\pm30$ & $2.37\pm0.01$ & $9390\pm30$ & $43.8\pm0.4$ & $0.238\pm0.001$ & $0.55\pm0.01$ & $7.38\pm0.01$ & $36.8\pm0.4$  \\
 \hline

\end{tabular} 
}
 \caption {Overview of characteristics calculated for the English PN, networks constructed from the various types of percolation experiments, link preserving randomization (MCMC), and networks grown by rejection sampling. $L$ is the number of links, $L_{0}$ the number of links in the giant component, $lr$ is the ratio of inter-layer over intra-layer links, $gc$ is the giant component size, $k_{max}$ is the maximum node degree, $CC$ is the average local clustering coefficient, $a$ is the assortativity coefficient, $d$ is the mean geodesic distance and $d_{max}$ is the network diameter. The networks grown by rejection sampling are those with word attachment ordered by word length (o) or at random (r) (cf. Sec. \ref{Growth}), and best fit core-periphery models that either match the size of the giant component and the number of links in it or additionally match the total number of links (cf. Sec. \ref{CP}). A maximum degree constraint with $k_\textrm{max}=20$ and $\nu=0.1$ was used. Averages over randomized ensembles are carried out by averaging over at least 10 ensemble members.}
\label{tab.1}
\end{center}
\end{table*}

A first reference case (which we refer to as type 0) to explore the organisation of words in real word repositories is given by models in which words are occupied at random. An appropriate model that respects the word length distribution might consider the union of subspaces $P_l\subset L_l$ constructed such that $H(l)$ unique words are chosen uniformly at random from $L_l$ to make up $P_l$. The pseudolexicon constructed this way is associated with a phonetic network which is analysed in Table \ref{tab.1}. In principle, the organisation of artificial words in $S$ is similar to English. One finds a giant component, lexical islands, and an overwhelming majority of hermit words. The presence of the giant component for words of length $l\leq 4$ is consistent with the Bethe estimates for the percolation thresholds, in fact layers are clearly supercritical for words up to length three and clearly subcritical for larger $l$. Hence, every artificial PN assembled at random in this way will have a ``core'' of densely occupied shorter layers that enable the formation of a giant component, cf. also the word length distributions restricted to words within the giant component shown in Figure \ref{fig.n1}. However, the contrast in quantitative comparisons to the real PN is striking: Our type 0 percolation experiments result in a by far smaller giant component and more than a factor of 10 less links than observed in the data.

Selecting words uniformly at random in the layers $L_l$ assumes building phoneme sequences by sampling phonemes uniformly at random from the alphabet $\cal P$. In real language, however, phoneme usage is not uniform, but highly skewed \cite{siew2013}. The above percolation experiments can be modified to account for such skewed phoneme frequencies. Instead of sampling words uniformly at random from layers, we construct $H(l)$ unique phoneme sequences of length $l$ for each layer $l$ by sampling phonemes from the phoneme frequency distribution determined for English. The resulting type 1 networks are analysed in table \ref{tab.1}, and again agreement with the general structure of the English PN can be stated. Quantitative comparisons yield slightly larger giant components and link counts in the artificial networks of type 1 compared to type 0.

Real language incorporates correlations between phonemes at other levels \cite{gruenenfelder2009}. Important among them are, e.g., consonant-vowel co-occurrence patterns in word production. To include such correlations, we develop a third type of percolation experiments (labelled type 2) in which phonemes are sampled from the real phoneme frequency distribution and empirically determined phoneme-phoneme correlations are respected when constructing artificial words, again in such a way that the resulting ensembles follow exactly the same word length distribution as the empirical data. Comparisons of artificial PNs pertaining to these ensembles are shown in table \ref{tab.1}, again noting a closer match in giant component size and link counts with the English PN. Another important observation is that (as noted in the experiments of \cite{gruenenfelder2009}) all of the artificial networks are marked by high clustering, high assortativity coefficient and small average (chemical) distances, but quantitative comparisons do not yield a good match within error bounds.

\begin{figure}[tbp]
 \begin{center}
\includegraphics[width=.6\textwidth]{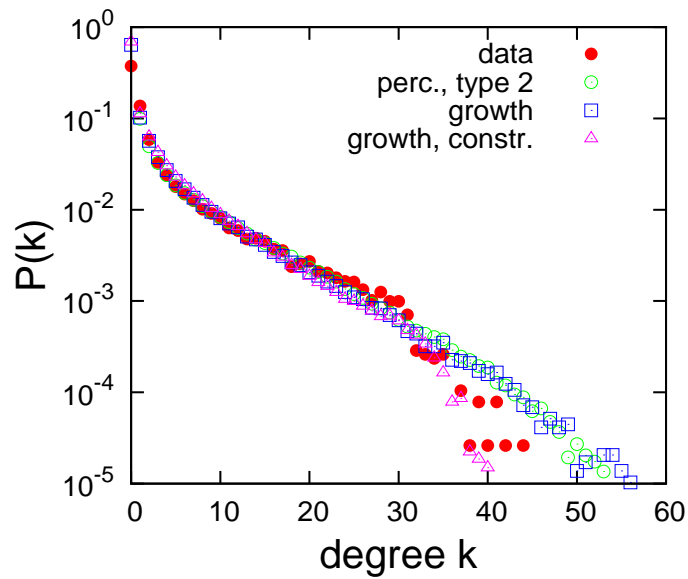}
 \caption {Degree distributions of the real PN and of artificial PNs, see legend. Data points for simulation experiments are averaged over at least 10 configurations.}
\label{fig.n2}
\end{center}
 \end{figure}

It is, however, interesting to note that the low degree region ($k<30$) of the degree distributions of all percolation experiments gives a very good match with the empirical data, in fact with the region that has previously been used to estimate a power-law dependence \cite{arbesman2010}. This is also the case with type 2 percolation, but even though these artificial PNs have significantly less links than the English PN already a heavier tail than observed for the real data is found. These observations support two conclusions: first, the power-law like region of the degree distribution results from the structure of the constraining space and from the decreasing word occupation density with word length so that no recourse to additional explanations is required (i.e. preferential attachment, as suggested in \cite{vitevitch2008}). Second, the English PN is characterised by a maximum degree cut-off such that words with large numbers of neighbours are suppressed in comparison to random sampling. As random sampling does not exhibit a similar cut-off, this cut-off does not result from constraints in the underlying space as speculated in \cite{gruenenfelder2009}, but must be caused by an additional constraining influence in word repertoire formation.

One obvious way to continue modelling artificial word repertoires is by including higher order correlations, for instance by fitting higher order Markov processes to the empirical data. Such an approach will naturally lead to a better fit of network metrics, but offers relatively little explanatory power. Instead, we note that all percolation experiments yield giant components and link counts much lower than measured for the English PN -- a consequence of the aforementioned higher order correlations in phonetic transcriptions of words. We next attempt to disentangle these observations of link count and component size. Are larger than expected component sizes just a consequence of larger link counts or do they represent additional peculiarities of the English PN?

\subsection{Randomization and Markov Chain Monte Carlo methods}
\label{sec:mcmc}

A standard way to explore peculiarities of a given network is via
randomization procedures that perform Monte Carlo steps designed to
destroy correlations in the given network's architecture while
enforcing constraints \cite{newman2010,snijders2002}. The method allows for the construction of ensembles of networks which preserve key characteristics of a given network, but are ``random'' in every other respect. Randomized networks constructed from such Markov Chain methods provide an important null model for network analysis; comparing a given network to such suitably randomized ensembles allows the identification of peculiarities of network structure. For this purpose one best proceeds by first randomizing network structures subject to the lowest level constraint, subsequently adding constraints in the course of the analysis then allows increasingly more sophisticated insights into higher order correlations peculiar to the given network data, see, e.g., \cite{milo2002} for applications of randomization techniques in the context of motif detection.

In this section we explore Monte Carlo schemes that preserve lowest
level network statistics while also enforcing spatial constraints,
phoneme statistics, and word length distributions typical to the
English PN. Proceeding systematically in this vein, one would first address the question of exploring the architecture of networks that have the same number of nodes as the given network, but are otherwise random (except for the above mentioned phoneme statistics, word lengths and spatial constraints). To construct such an ensemble, one could proceed as follows: (i) randomly select a word $w$ in the current repertoire and remove it. (ii) Generate an artificial word  $w\prime$ of the same length as the word $w$ in such a way that phonemes are drawn from the phoneme frequency distribution corresponding to the original repertoire and such that phoneme-phoneme correlations as observed in the original repertoire are on average conserved. (iii) Accept the word $w$ if it does not coincide with another word already in the repertoire. If $w\prime$ is not accepted, $w$ is inserted again and we proceed with step (i) selecting another word $w$ at random. It is straightforward to see that the suggested procedure will asymptotically result in ensembles as the ones discussed as ``type 2 percolation'' in the previous section, i.e. essentially random allocations of occupied and empty nodes in word space.

Increasing the order of preserved network statistics, we next analyse ensembles of networks that have the same number of nodes and links and phoneme and word lengths statistics as the given English PN, but are otherwise random. For this purpose we proceed as suggested above, but modify the acceptance step (iii) such that suggested words $w\prime$ are only accepted if the new suggested word does not already exist in the repertoire and if it has the same degree as the previous word $w$.

\begin{figure}[tbp]
 \begin{center}
\includegraphics[width=.8\textwidth]{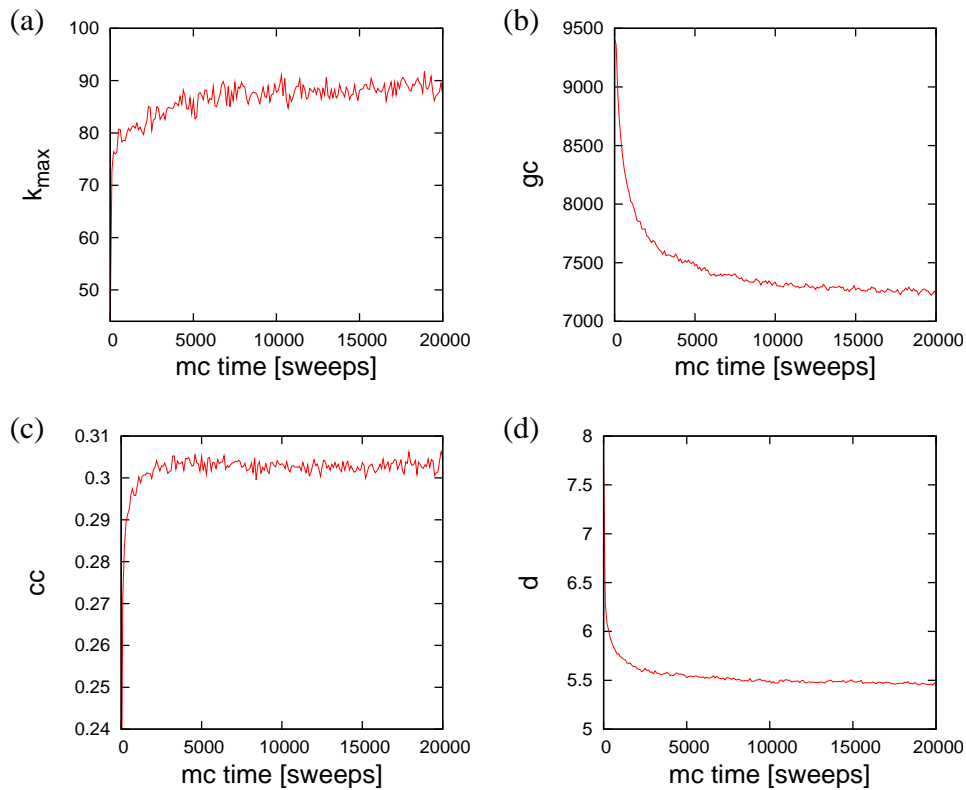}
 \caption {Change in (a) maximum degree, (b) size of the giant component, (c) clustering coefficient, and (d) average shortest pathlengths with the number of randomization steps (measured in multiples of the network's size). The initial network is the English PN and then randomization steps that preserve the network's link density are performed. The curves correspond to averages over ten independent runs. }
\label{fig.nn1}
\end{center}
 \end{figure}

Figure \ref{fig.nn1} shows the change in some network characteristics versus the number of randomization steps performed, measured in multiples of system size. The initial points in each panel reflect the data measured for the English PN, the following evolution shows how certain correlations peculiar to the English PN are gradually destroyed as the network is randomized. The simulation data illustrate that the method is computationally quite demanding, only after around 10.000 randomization sweeps an equilibrium is reached. Next, characteristics of the ensemble of constrained networks with the same number of nodes and links as the English PN can be measured and averaged over the following 10.000 Monte Carlo updates.

Averages over some network measures that characterise the randomized ensemble are given in Table \ref{tab.1}. Results essentially confirm earlier observations from percolation experiments: The English PN is found to have a substantially smaller tail of highly connected words, is less cliquish and has stronger assortative mixing by degree. Further, we observe that the English  PN has a significantly larger and less densely connected giant component than expected. Larger than expected average path lengths and diameters appear as a consequence of the larger size of the giant component.

While confirming earlier results, comparisons to the ensemble of random PNs with the same link density also allows for additional conclusions. We find that the reduced size of the giant component and enhanced clustering and assortative mixing coefficients observed in the percolation experiments are not artifacts of reduced link counts, but hint at additional features characteristic of the English PN itself.

Our first steps of a randomization analysis can be extended by fixing additional constraints which would allow deeper analysis of factors that contribute to the architecture of the English PN. We have undertaken some preliminary experiments in this direction, but
computational costs become increasingly onerous as the rejection
rate increases and the calculation of global network statistics at
each Monte Carlo step can become very demanding. Instead of pursuing randomization ideas subject to additional constraints, a comparison to ensembles of grown networks is a more flexible approach that sheds light on possible non-equilibrium features of repertoire formation. In particular, we can use it to explore what constraints are required in models of network growth to match important statistics of the English PN.

What we have observed so far is that without enforcing additional constraints randomization-based null models can neither satisfactorily reproduce the size of the giant component nor core-periphery features of the empirical data set.  These larger than expected connectivity properties hint at a process of repertoire formation in which words are assembled over time in such a way that preferentially such new words that connect to older words are added. Alternatively, one can interpret this as word assembly in a way that new words are preferentially formed by slightly modifying already existing word forms, as in word derivation \cite{aitchison2012}. How likely is it that such a process could form PNs that are in quantitative agreement with the real data? Can the real data help to estimate constraints in such a process of word assembly? These are the questions we are going to explore in the remainder of the paper.

\section{Growing repertoires by rejection sampling}
\label{Growth}

In this section we explore processes of word repertoire formation over time, i.e. growth models for the phonetic networks corresponding to the artificial repertoires. We start with the observation of the differences in the size of the giant components reported in the previous section. This motivates a first type of growth process, described in subsection \ref{G1}, which aims at generating artificial networks with similar size of the giant component as the English PN. These models are then refined in subsection \ref{CP}, in which we develop models that match further connectivity properties of the real data.  

Before proceeding it is worth pointing out that all our models cannot capture the real evolution of the English language over time. The main reason for this is that the framework lacks any realistic assumption about the role played by the influence of external languages and about the rate of language change. Nonetheless, the following section will investigate if models of growing networks can reproduce features of the English PN, and will explore what constraints in word assembly are likely to have played a role in repertoire formation.

\subsection{Growth models}
\label{G1}

\begin{figure*}[tbp]
 \begin{center}
\includegraphics[width=.9\textwidth]{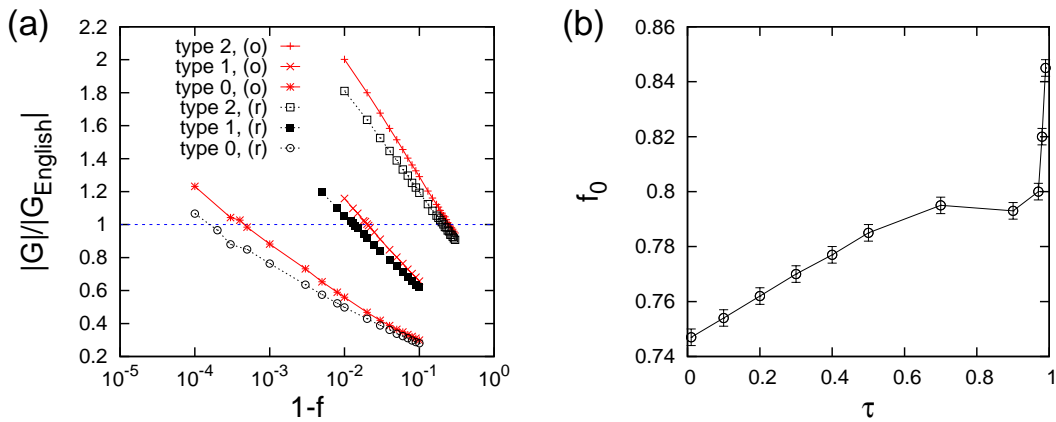}
 \caption {(a) Dependence of the relative size of the giant component of artificially grown word ensembles on the rejection probability $f$ for both word length ordered and random attachment for artificially constructed words according to the percolation 0,1, or 2 rules. Data for degree constrained ensembles are not shown as data points are virtually identical. Data points represent averages over at least 10 networks. (b) Dependence of the probability $f$ that allows to reach the same size of the giant component as in the real English PN on the order of attachment of words as measured by the parameter $\tau$, cf. text. }
\label{fig.n3}
\end{center}
 \end{figure*}


Consider a process of word assembly in which new words are added by suggesting new randomly sampled words of lengths drawn from a given word length distribution $H(l)$ and rejecting them with some probability
\begin{align}
r=f +(1-f) p_k
\label{E1}
\end{align}
according to an acceptance criterion. The severity of the application of this criterion is tunable by a parameter $f$ and the probability $p_k$ implements additional degree constraints explained below. New artificial candidate words can be generated according to the type 0, 1, or 2 percolation models detailed in the previous section, i.e. by uniform sampling from the set of all phonemes, sampling from the real phoneme distribution or additionally respecting phoneme-phoneme correlations.

To generate artificial ensembles that reach the same connectivity properties as the English PN we accept new words with probability $r$ if they connect to at least one old word and reject them otherwise. In both cases a new artificial word is suggested until exactly as many words of length $l$ have been accepted as present in the word length distribution $H(l)$ of the English repertoire. As the previous section has demonstrated the presence of an additional degree constraint in the English PN the second factor in Eq. (\ref{E1}) serves to suppress nodes of large degree. For a suggested word $w$ we set
\begin{align}
p_k= \exp(\nu (k_\textrm{max}-k_w)) \prod_{n \in {\cal N}_w} \exp(\nu (k_\textrm{max}-k_n)),
\label{E2}
\end{align}
where ${\cal N}_w$ is the set of all neighbours of $w$, $k_w$ the degree of $w$, and $k_{\textrm{max}}$ and $\nu$ parametrise the cut-off behaviour for large degrees. Note that the major difference to established uniform attachment models \cite{callaway2001} is the constraint of the underlying space. Different from \cite{callaway2001} new nodes in our model can have different degrees, depending on their location in the space $S$. The degree constraint expressed in Eq. (\ref{E2}) is essentially implemented in such a way that new nodes are likely to be rejected if any of their neighbours would reach too large a degree by the addition of the new node. Using the above framework additional constraints can easily be implemented as additional terms to Eq. (\ref{E1}), some of which we will discuss in more detail later.

By showing the dependence of average sizes of giant components of grown ensembles on the acceptance probability Figure \ref{fig.n3}(a) summarises some first simulation experiments. Stronger acceptance constraints (i.e. larger $f$) allows to construct networks with larger giant components and by estimating the crossing of simulation data with the line $|G|/|G_{\textrm{English}}|=1$ one can find values for the rejection probability $f$ such that the resultant ensembles will match the size of the giant component of the English PN. Comparing this crossing value $f_0$ for different growth procedures allows estimates about the relative likelihood of reproducing realistic features of the English PN using these procedures. For instance, with $f_0^{\text{(type 0)}} \approx 1-10^{-4}$ generating realistic ensembles by uniform attachment proves excruciatingly difficult and even ignoring phoneme-phoneme correlations by only sampling from the real phoneme frequency distribution one still has $f_0^{\text{(type 1)}}\approx 1-10^{-2}$, while inserting phoneme correlations yields  $f_0^{\text{(type 2)}}\approx 0.75$. As one would naturally expect, since $f_0^{\text{(type 0)}}>f_0^{\text{(type 1)}}>f_0^{\text{(type 2)}}$ we observe that growing ensembles by suggesting new words that include more realistic phoneme statistics provides a more likely explanation of the real data. For this reason, and as we aim to construct word repertoires that respect lower order correlations in the real data we proceed with word generation method type 2 in all experiments presented below.

Words can be attached in different order and the order of attachment will generally influence the structure of the generated network. Panel (a) of figure \ref{fig.n3} also compares attachment of words in random order (black symbols) and attachment ordered by word length, starting from the shortest words (red symbols). One generally finds $f_0^{\text{(ordered)}}<f_0^{\text{(random)}}$, i.e. attachment ordered by word length gives a more likely explanation of the data than random attachment. Similar experiments can also be carried out using other attachment criteria (e.g. accepting words if they connect to the giant component and rejecting them with probability $f$ otherwise) or by determining crossing probabilities for other network statistics (e.g. the number of links). We have tested some of these alternatives and found similar qualitative results, but different numerical values of the estimated crossing probabilities $f_0$. Link attachment was chosen as the most suitable starting point for further exploration below.

In panel (b) of Figure  \ref{fig.n3} the attachment order is explored more systematically. For this purpose we introduce an additional attachment order parameter $\tau$ and construct lists of word lengths according to which words are attached in the following way. We start by setting $H^\prime=H$. For a given place in the list, say $t$, a word will be allocated the smallest $l$ bin of $H^\prime(l)$ which has not been exhausted yet, i.e. for which $H^\prime(l)>0$. Then, we continue increasing $l$ with probability $\tau$. If a value of $l$ was generated this way for which $H^\prime(l)=0$ we select the closest smaller value of $l$ for which $H^\prime(l)>0$. Having thus determined the word length of the word which will be attached at step $t$ we decrease $H^\prime(l)$ by one and repeat the process until $t=N$ has been reached. The parameter $\tau$ is thus a measure of the order of attachment of new words. For $\tau=0$ words are attached ordered exactly by word length, starting with the shortest words. For small $\tau$ attachment is generally ordered by word length, but there is a small chance that slightly longer words might sometimes precede shorter words. If $\tau=1$ words are attached in reverse order, i.e. the longest words first and the shortest ones last. We find that in between for $\tau\approx 0.9$ the attachment order is approximately random.

Systematically varying $\tau$ we find that attachment orders that attach shorter words first are generally more likely explanations of the data than other word orders, cf. the monotonic trend in Figure \ref{fig.n3}b. This is compatible with the hypothesis in psycholinguistics that the mental lexicon is built starting preferentially with shorter words, that are also more frequent \cite{storkel2004,hills2010,aitchison2012,siew2013}. Consequently, in all experiments presented below we will assume almost perfectly ordered attachment of words only allowing for a small amount of disorder $\tau=0.01$ to ensure for robustness.

Growing a network by attaching nodes over time introduces correlations which can alter the degree distribution in comparison to models that allocate links randomly in a fixed set of nodes. Figure \ref{fig.n2} also compares the degree distributions of the grown ensembles with the real data. Again, as in the percolation experiments before, we note that the low degree part of the distribution is well matched by the networks associated with grown word ensembles, but growth also induces a heavier tail for large degrees. A good match with the data for the English PN can be obtained if one sets $k_{\text{max}}=25$ and $\nu=0.1$, cf. the data for ensembles grown with constraints in Figure \ref{fig.n2}. Ensembles grown with constraints to match the size of the giant component of the English PN also have the same distribution of word lengths in the giant component as the English PN, cf. Figure \ref{fig.n1}.

\begin{figure}[tbp]
 \begin{center}
\includegraphics[width=.65\textwidth]{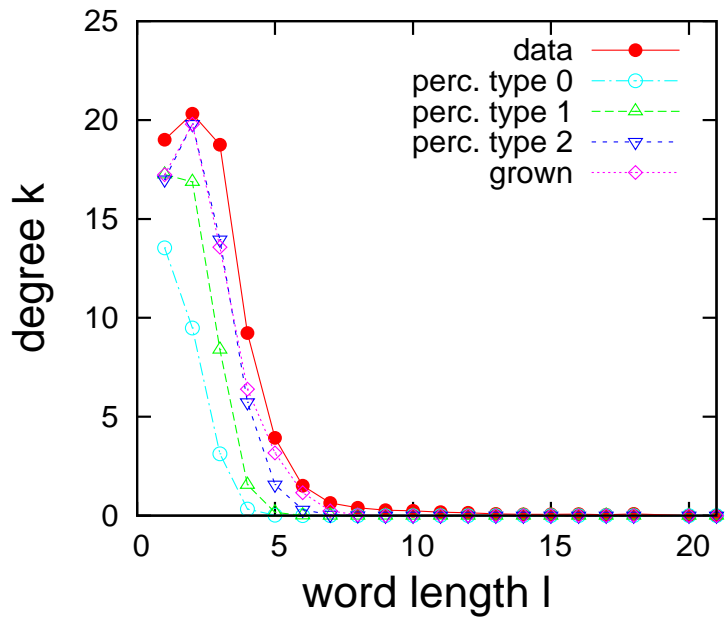}
 \caption {Average degree vs word length for the English phonetic network and various null models. For the simulation data averages over at least 10 independent runs have been taken.}
\label{fig.n4}
\end{center}
 \end{figure}

Further analysis of the networks belonging to the grown ensembles is presented in table \ref{tab.1}. Most prominently, one notes that all such networks have less links than the English PN, but also other network statistics show quantitative deviations whilst generally confirming qualitative observations about large clustering, high assortativity and the small world character of artificial PNs. One is thus lead to wonder which links are missing in the grown artificial PNs in comparison to the English PN. Figure \ref{fig.n4}  addresses this question by plotting average degree vs word length for the English PN and null models constructed by percolation or growth. As one would expect from our earlier arguments about percolation thresholds and the structure of the underlying space, links are densely concentrated on shorter words, leading to a clear core-periphery structure of the networks. Average nodes corresponding to word lengths larger than 5 or 6 have hardly any network neighbours, whilst nodes belonging to shorter words are very densely connected, having around 20 neighbours on average. Furthermore, a comparison between the null models and the data shows that mostly links within the core are not captured by our modelling yet. In particular, comparing percolation type 2 experiments and ensemble growth one notices that the link attachment constraint mostly adds links for words longer than four phonemes while leaving the average link count for shorter words almost unaltered. This is the case because short words almost always have a neighbour and will thus be preferentially accepted, such that the link attachment constraint only becomes effective in adding connections for longer words. Consequently, an improved null model will have to account for more links for short words. In the next section, we propose a family of core-periphery models aiming to generate artificial PNs with giant components and link counts matching the English PN.

\subsection{Core periphery models}
\label{CP}

The main purpose of our core periphery (CP) models is to account for missing links in null models for short words relative to the data. Hence, we define these models by the iteration of the following steps: (i) construct an artificial (non duplicated) word according to the type 2 process, (ii) reject this word with probability
\begin{align}
r=f + (1-f) p_k + (1-f)(1-p_k) C(w,k_w),
\label{E3}
\end{align}
where, as in previous experiments, the factor $f$ gives the tunable rejection probability, $p_k$ models the maximum degree constraint according to equation (\ref{E1}), and the additional factor $C(l_w,k_w)$ models a core-periphery constraint. For a suggested word $w$ of length $l_w$ and degree $k_w$ we set
\begin{align}
C(w,k_w) =
  \begin{cases}
   \delta & \text{if } l_w<W_C \text{ and } k_w<m_C \\
   0       & \text{if }l_w\geq W_C
  \end{cases},
\end{align}
in which $W_C$ models the core size (in terms of word length) and $m_C$ \footnote {Fractional values of $m_C$ are interpreted probabilistically, i.e. with prob. $m_C-\lfloor m_C \rfloor$ we set the value to $\lceil m_C \rceil$ and to $\lfloor m_C \rfloor$ otherwise.} allows to tune the density of links in the core and $\delta$ is the strictness of the application of the core criterion for short words. Simply put, we suppress  short words that do not have enough links, in order to counter the effect reported in Figure \ref{fig.n4}. Since there are rare configurations in which the core criterion cannot be exactly met, we set $\delta=0.99$ in all following experiments. The modified procedure allows us to tune the number of links in the core. Additionally, the parameter $W_C$ allows to vary the size of the core, whereas $m_C$ allows to tune the link density within the core. Comparison to the plot of average degrees versus word length in Figure \ref{fig.n4} suggests that $4 \leq W_C \leq 6$ for English.

\begin{figure*}[tbp]
 \begin{center}
\includegraphics[width=.9\textwidth]{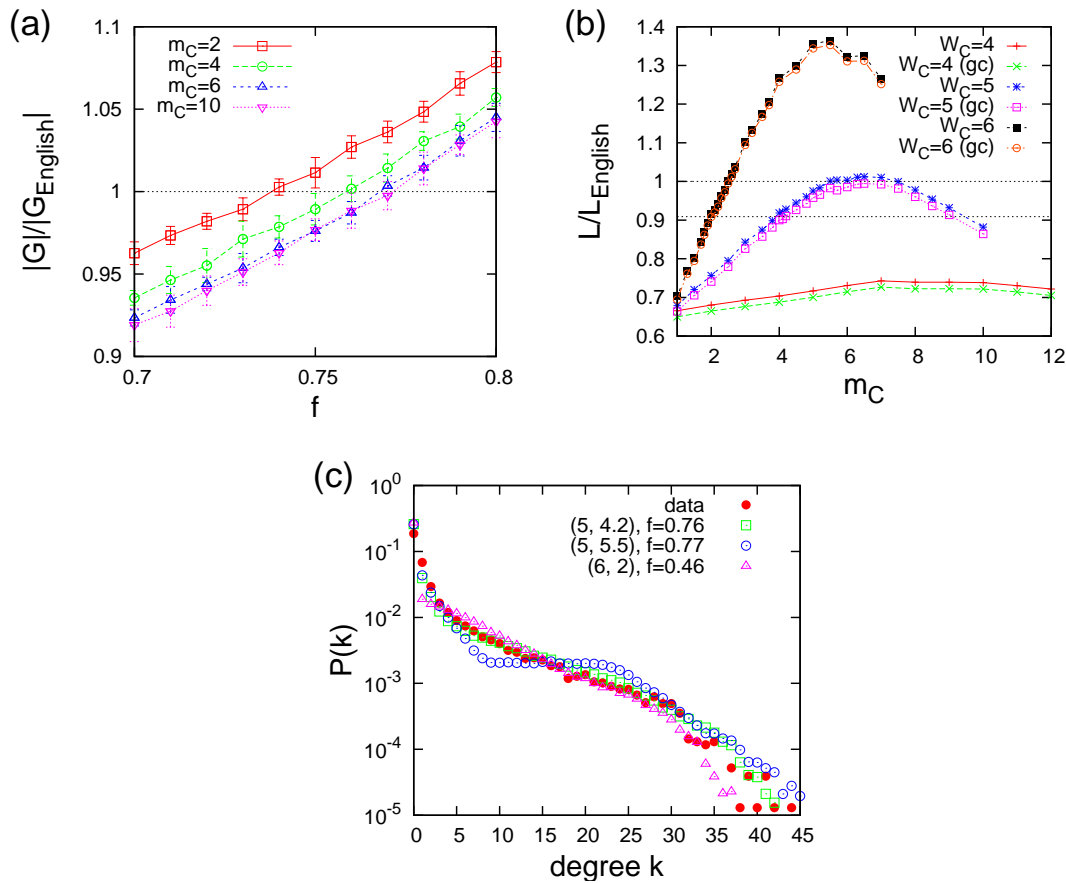}

 \caption {(a) Relative size of the giant component vs rejection probability for CP models with $W_C=5$ and several choices of $m_C$. (b) Analysis of core periphery (CP) models. We note that for small core sizes $W_C\leq 4$ it is not possible to build networks with the same number of links as the real data, the same is true also for $W_C=7$ irrespective of $m_C$. For $W_C=5$ and $W_C=6$, there are generally two intersection points. The following CP networks with same size of giant component and same number of links in the giant component as the real network emerge: $(W_C,m_C)=(5,4.2), (5, 9)$ and $(6,2)$. (c) Degree distributions of candidate CP networks compared to empirical data. If the core size is too large, deviations for low degrees occur (see (6,2)). Likewise, if $m_C$ is too large relative to the core, nodes with $k<m_C$ are underestimated compared to the data. This only leaves the low $m_C$ intersection point for core $W_C=5$ networks as reasonable candidate models which replicate the size of the giant component, number of links, and the degree distribution of the English PN. Results represent averages over at least 10 configurations.}
\label{fig.n5}
\end{center}
 \end{figure*}

To explore which core-periphery models give a good description of the English PN we proceed as before. For each combination $(W_C, m_C)$ a rejection probability $f$ can be determined such that the respective CP network matches the size of the giant component of the English PN, cf. panel (a) of Figure \ref{fig.n5} in which data for $W_C=5$ and various values of $m_C$ are analysed. Once this probability $f$ has been determined link counts for links within and outside of the giant component can be compared for various core densities $m_C$, see panel (b) of Figure \ref{fig.n5} which compares link count vs $m_C$ dependencies for cores of various sizes. One notes that generally almost all links belong to the giant component. For better comparisons of properties of the giant component we determine intersection points of link counts in the giant component. As one would expect, link numbers increase at first when core connectivity $m_C$ is increased. When large values of $m_C$ are chosen the requirement for new nodes within the core to be accepted becomes very demanding and since the core requirement was implemented probabilistically increasingly more nodes are accepted without fulfilling it. This explains a reversal in trend, such that for each core size two intersection points at which CP networks of a certain core size match the number of links within the giant component of the English PN can be identified. This, however, is only the case if the core size $W_C$ is large enough. Small cores are composed of too few words to allow for the addition of enough links to reach the required link number for comparison with the English PN, this is for instance the case for $W_C=4$, cf. data in panel (b) of Figure  \ref{fig.n5}. Following this argument four candidate parameter sets for comparison to the English PN are found, i.e. the low and high $m_C$ intersection points for $W_C=5$ and $W_C=6$ (Figure  \ref{fig.n5}).

When comparing the degree distributions of these networks to the English PN it becomes apparent that only the low $m_C$ intersection point for $W_C=5$ gives a reasonable match. If core connectivity is chosen too high relative to core size, too many high degree core nodes are generated while low degree nodes are underrepresented. The large degree cut-off then results in a degree distribution with a plateau (open circles in panel (c) of Figure \ref{fig.n5}). Similarly, if the core size is chosen too large (i.e. $W_C=6$ or larger) not enough nodes of degree one or two are generated to allow for a good comparison to the English PN. Hence, only one ensemble of CP networks is identified which matches the size of the giant component, number of links in the giant component, and gives a good fit of the degree distribution of the English PN. Further analysis for this ensemble, constructed with $(W_C,m_C)=(5,4.2)$ and $f=0.76$, is presented in table \ref{tab.1}. Comparing network statistics with the English PN, there is a large discrepancy between link counts within and outside of the giant component. Like for all percolation type networks also for all the grown network ensembles very few links connect nodes that do not belong to the giant component. Hence, comparisons of properties of the entire network are not yet reasonable, since a very large fraction of all nodes has significantly less connections than in the English PN.

The last observation motivates us to introduce a last set of null models in which the number of links in- and outside of the giant component can be tuned. To define such a variant of CP networks we add another term to Eq. (\ref{E3}) which now becomes
\begin{multline}
r=f + (1-f) p_k + (1-f)(1-p_k) C(w,k_w)+ \\+ (1-f)(1-p_k) (1-C(w,k_w)) R (w),
\label{E4}
\end{multline}
where
\begin{align}
R(w) =
  \begin{cases}
   \epsilon & \text{if } w\in G_t \\
   0       & \text{otherwise }
  \end{cases}
\label{E41}
\end{align}
is an additional term that accounts for the rejection of new nodes if they link to the largest component $G_t$ at iteration $t$ of the assembly process. By tuning the probability $\epsilon$ in Eq. (\ref{E41}) we can construct ensembles of CP networks with relative fractions of links in the giant component lower than found in the models characterized by Eq. (\ref{E3}) which is retrieved for $\epsilon=0$.

\begin{figure*}[tbp]
 \begin{center}
\includegraphics[width=.9\textwidth]{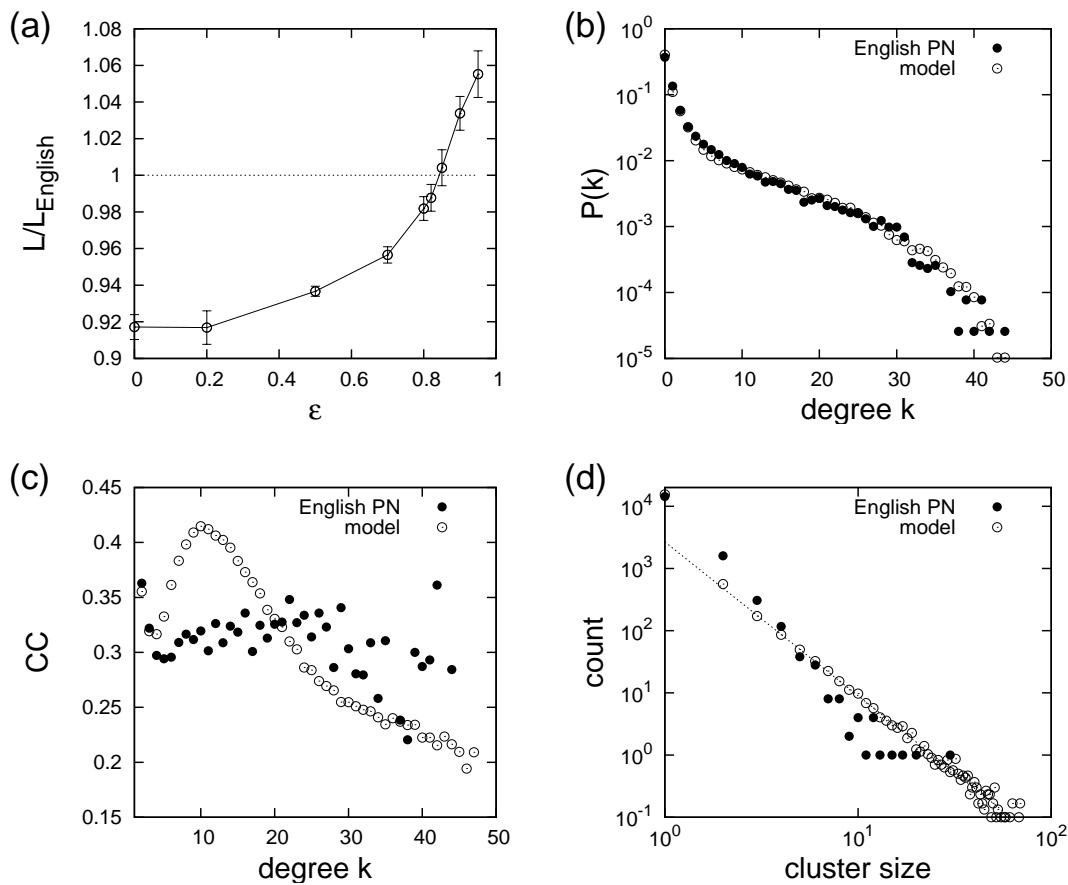}

 \caption {(a) Ratio of link counts of CP networks and the English PN vs. the parameter $\epsilon$. The link count inside of and outside of the giant component are met for $\epsilon \approx 0.82$ (and $W_C=5, m_C=4.4, f=0.962$). (b) Comparison of the degree distribution of the above network and the English PN. (c) Local clustering coefficient vs degree for the English PN and the CP networks. (d) Comparison of the component size histograms for the English PN and our model. Data points represent averages over at least 10 configurations.}
\label{fig.n6}
\end{center}
 \end{figure*}

In panel (a) of Figure \ref{fig.n6} the $\epsilon$-dependence of CP networks with $W_C=5$ is explored. Data points in the figure are obtained in the following way. For fixed value of the parameter $\epsilon$ and given core connectivity $m_C$ the parameter $f$ for which the ensemble reproduces the size of the giant component of the English PN is determined. As in the analysis of the model described by Eq. (\ref{E3}), varying $m_C$ the intersection points at which the networks give link counts within the giant component identical to the data in the English PN can be determined and average total link counts of these networks are than plotted. In this way we can identify the CP model which reproduces all criteria for a valid comparison to the English PN: These networks reproduce the size of the giant component, link counts within and outside of the giant component, and the degree distribution of the English PN (for the latter see panel (b)).

Relevant network statistics of this ensemble constructed with $(W_C,m_C)=(5,4.4)$, $\epsilon=0.82$, and $f=0.962$ are given in table \ref{tab.1}. As with all comparisons of null models along the way, we notice that whilst the null models suggest that phoneme networks should be highly clustered and highly assortative by degree, quantitative comparison yield that:
\begin{itemize}
\item [(i)] The English PN is significantly less cliquish than what would be expected from the null model, i.e. the null model predicts a clustering coefficient of $CC=0.238\pm 0.004$ whereas $CC=0.207$ for the English PN. In fact, quantitative comparisons of the dependence of clustering on degree can be carried out and show that lowly connected words in the English PN are part of less triangles than expected in the CP networks, whereas large degree nodes are part of more triangles than expected, cf. Figure \ref{fig.n6}(c). Due to peculiarities of the embedding space we cannot expect the $1/k$ dependence typical in preferential attachment models \cite{barabasi1999}.
\item [(ii)] The English PN has an assortativity coefficient significantly higher than expected, i.e. $a=0.55\pm.01$ is predicted by the null model and $a=0.707$ is found for English.
\item [(iii)] Average path length and diameter of the English PN are (roughly) compatible with the expectations from the null model, i.e. the null model predicts $d=7.38\pm 0.3$ and $d_\text{max}=36.8\pm 5.2$ whereas $d=7.71$ and $d_\text{max}=33$ for English.
\item [(iv)] Also the arrangement of links into intra- and inter-layer connections are roughly compatible between the null model and the English PN, we find that the ratio of intra- to inter layer links is $lr=2.37\pm 0.07$ for the null model, whereas $lr=2.46$ for English.
\item [(v)] The CP null model predicts significantly larger small clusters than found in English, see panel (d) of Figure \ref{fig.n6}. In fact, ignoring the topology of the underlying space, a model in which new nodes are accepted if they connect to old nodes with a certain probability, implements a preferential attachment mechanism as described by Barabasi and Albert for degrees \cite{barabasi1999} for cluster sizes. One thus expects a power law with exponent close to $3$ for the distribution of small clusters and, since node additions can join clusters, lower exponents in the presence of constraints from an embedding space \cite{newman2010}. 
\end{itemize}

\section{Conclusions and discussion}

The English phonological network represents a snapshot of the organisational patterns of word pronunciation in the human mental lexicon. In the present study we started by recognising that English words are effectively a subset of the set of all possible words formed by all possible combinations of phonemes. The latter, i.e. the set of all possible words endowed with the edit distance as a metric, defines a high dimensional discrete space which can be visualised as a stack of structured sets of words of given lengths (which we call \textit{layers}). Phonological networks are embedded into this space. We systematically explored how spatial characteristics influence word pronunciation patterns, thereby revealing characteristics of the English language.

Percolation experiments demonstrate that some features of the English PN are a consequence of the embedding space. Importantly, we find that the presence of a power-law like regime in the degree distribution arises also in pseudolexica constructed by random sampling, i.e. contrary to what was suggested in \cite{vitevitch2008}, no additional attachment mechanism like preferential attachment needs to be invoked as an explanation. Furthermore, our percolation models highlight the presence of a maximum degree constraint on the PN that is not a direct consequence of the embedding space. This finding suggests the presence of a maximum number of phonologically similar words that can be associated and stored, i.e. it points to a constraint of word confusability \cite{sadat2014} in word repertoire formation. However, percolation experiments cannot reproduce connectivity properties of the English PN. In fact, all PNs associated with artificial repertoires constructed via percolation have substantially smaller link counts and sizes of the giant component than the real PN. 

These insights can be refined by further comparisons to ensembles of networks with the same number of connections as the English PN. Randomization experiments along these lines point out that the smaller than expected sizes of the giant components generated by percolation-like experiments are not only a consequence of reduced link counts. We conclude that the rather large size of the giant component is a characteristic of word organisation. 

An explanation for this enhanced connectivity of the English PN is word repertoire formation through a process of constrained word assembly over time, in such a way that preferentially connected words are included. We systematically explore this idea by introducing a series of network growth models. We first focus on the sizes of giant components and consider models in which new words are rejected if not linked to older words. Quantitative analysis leads to three main conclusions. First, the growth models corroborate the findings of constraints on maximum degree in repertoire formation. Second, within the constraints of our models, word assembly ordered by word length is a likelier explanation of the data than random word addition, giving a quantitative basis to the hypothesis that language evolved from short to long words, similar to the language acquisition of children who tend to learn shorter words first \cite{aitchison2012,carlson2011,storkel2004,vitevitch2008}. Third, the analysis points towards a marked \textit{core-periphery structure} of the English PN, suggesting that in the earlier stages of repertoire formation, preferentially such short words which are similar to (or can be derived from) multiple existing words have been assembled to the language, as already suggested in the psycholinguistic literature \cite{siew2013}. This latter finding inspires the introduction of core periphery (CP) network models.

CP network models of repertoire formation can reproduce the size of the giant component and the number of links within the giant component of the English PN. The English PN, however, has a far larger number of edges between nodes in smaller clusters than predicted, motivating the introduction of a last type of CP networks with tunable link counts in- and outside of the giant component. These networks, finally, provide null models which retrieve the size of the giant component, link counts, and the degree distribution of the English PN, and hence a systematic comparison of higher order correlations in network structure becomes possible. Several additional features of network organisation are well-represented by expectations from these reference CP models: diameters and distances fall within the error bounds of prediction and the link organisation in and between the layers are in good agreement. In contrast to previous work, however, these comparisons point out that the English PN is less cliquish and more assortative than expected. The first is a feature that might point to further constraints in repertoire formation. Similar to the degree constraint, suppression of triangles might point towards a mechanism of word formation that under-represents words that are too similar to others. 

Whilst our study has highlighted and explored some constraints likely at play in repertoire formation, other features of the English PN are not adequately captured or well enough understood by the models we presented. This applies to detailed patterns of cliquishness vs degree, the detailed statistics of smaller components or a better understanding of the very high assortative mixing by degree of the English PN. In the spirit of the first empirical analysis of \cite{arbesman2010} for Spanish, Hawaiian, Mandarin and Basque, our null models also enable a detailed comparison with other languages in future work. Are the same assembly mechanisms at play in all languages? How can differences in assembly be explained or related to cultural peculiarities? These are questions beyond the scope of a physics approach, but the methodology suggested here might enable linguists to explore them quantitatively.

\section{Acknowledgements}
The authors acknowledge the IRIDIS High Performance Computing Facility, and associated support services at the University of Southampton, in the completion of this work. MS was supported by an EPSRC grant (EP/G03690X/1). We also wish to acknowledge reviewer comments that helped to improve the structure of the paper.

\section{References}

\bibliography{bibliography}
\end{document}